\documentclass{article} 
\usepackage{iclr2022_conference,times}

\usepackage{amsmath,amsfonts,bm}

\def\eqref#1{equation~\ref{#1}}

\def\1{\bm{1}}

\def\vb{{\bm{b}}}

\DeclareMathAlphabet{\mathsfit}{\encodingdefault}{\sfdefault}{m}{sl}
\SetMathAlphabet{\mathsfit}{bold}{\encodingdefault}{\sfdefault}{bx}{n}

\usepackage{hyperref}
\usepackage{url}
\usepackage{graphicx}
\usepackage{placeins}

\usepackage{physics}

\def\RefImageB{\ensuremath{I_B}}
\def\RefImageA{\ensuremath{I_A}}

\def\CLIPEmbedImage{\ensuremath{E_I}}

\def\RefWplus{\ensuremath{w^{\text{ref}}}}
\def\RefAccrossDomainShift{\ensuremath{v^\text{ref}}}
\def\SampleAccrossDomainShift{\ensuremath{v^\text{samp}}}

\title{ Mind the Gap: Domain Gap Control for Single Shot Domain Adaptation for Generative Adversarial Networks}

\author{Peihao Zhu\\
KAUST
\And
Rameen Abdal\\
KAUST
\And
John Femiani\\
Miami University
\And
Peter Wonka\\
KAUST
}

\iclrfinalcopy 
\begin{document}

\maketitle

\begin{figure}[h]
    \centering
    \includegraphics[width=0.96\textwidth]{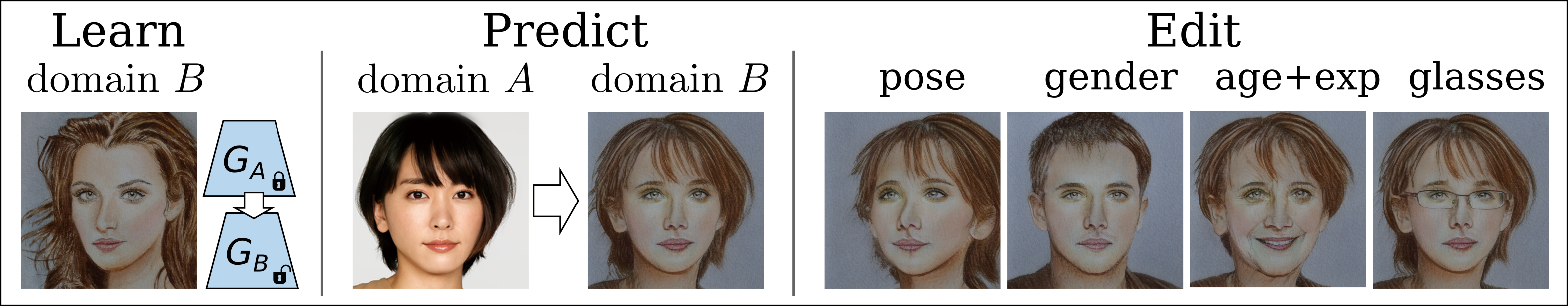}
    \caption{One-shot domain adaptation:   (left) a single reference image from domain $B$  is used to refine a GAN $G_A$ to learn $G_B$; (center) every image in domain $A$ has an analog in domain $B$ that shares a latent code and many salient attributes; (right) because salient attributes are preserved in the new domain, many latent-edits are meaningful in the new domain.}
    \label{fig:teaser}
\end{figure}

\begin{abstract}
We present a new method for one shot domain adaptation. The input to our method is a trained GAN that can produce images in domain $A$ and a single reference image $I_{B}$ from domain $B$. The proposed algorithm can translate any output of the trained GAN from domain $A$ to domain $B$. There are two main advantages of our method compared to the current state of the art: First, our solution achieves higher visual quality, e.g. by noticeably reducing overfitting. Second, our solution allows for more degrees of freedom to control the domain gap, i.e. what aspects of the image $I_{B}$ are used to define the domain $B$.
Technically, we realize the new method by building on a pre-trained StyleGAN generator as GAN and a pre-trained CLIP model for representing the domain gap. We propose several new regularizers for controlling the domain gap to optimize the weights of the pre-trained StyleGAN generator so that it will output images in domain $B$ instead of domain $A$. The regularizers prevent the optimization from taking on too many attributes of the single reference image. Our results show significant visual improvements over the state of the art as well as multiple applications that highlight improved control.

\end{abstract}

\section{Introduction}

We propose a new method for domain adaptation based on a single target image.
As shown in Fig.~\ref{fig:teaser}, given a trained GAN for domain $A$, and a single image \RefImageB{} from domain $B$,
our approach learns to find a corresponding image in domain $B$ for any image in domain $A$. We can achieve this by fine-tuning the GAN for domain $A$ to obtain a second GAN that generates images in domain $B$. The two GANs share a latent space so that a single latent code will generate two corresponding images, one in domain $A$ and one in domain $B$.
The main selling point of our method is that it achieves superior quality than the state of the art in single shot domain adaption. 
Our method is computationally lightweight and only takes a few minutes on a single GPU, so that it can be widely applied.

In order to do this, we leverage multiple existing components, including two excellent pre-trained networks: First, we use StyleGAN2~\citep{Karras2019stylegan2} as a pre-trained GAN. A follow-up version has been published on arXiv~\citep{Karras2021}, but the code only became available after we finished all experiments. Second, we use a pre-trained network for image embedding, CLIP~\citep{radford2021learning}, to encode images as vectors.
Third, we use the pioneering idea of StyleGAN-NADA~\citep{gal2021stylegannada}, which builds upon StyleCLIP~\citep{patashnik2021styleclip}, to encode a domain gap (or domain shift) as vector in CLIP embedding space. Fourth, we leverage II2S~\citep{zhu2020improved} as GAN embedding method to transfer image $I_{B}$ into domain $A$ to obtain a better estimation of the domain gap.

Even though the visual quality of StyleGAN-NADA is already impressive when used as a single image domain adaption method, we identified multiple technical issues that can be improved to achieve another large jump in visual quality. First, and most importantly, StyleGAN-NADA was designed for zero-shot domain adaptation, and does not have a good solution to model the domain gap based on a single example image. Their reference implementation models the domain gap as a vector from the average image in domain $A$ to the given image \RefImageB{} in CLIP embedding space. However, this leads to overfitting in practice and the transfer results lose  attributes of the input images, so that input images from domain $A$ get mapped to images that are all too similar to \RefImageB{} in domain $B$. We identify a better solution to this problem. In fact, the domain gap should be modeled as a vector from the image \RefImageB{} to its analog in domain $A$, so that the image in domain $A$ shares salient within-domain attributes with the reference image. We therefore need to solve an inverse $B$-to-$A$ domain-transfer problem, which we propose to tackle using the state-of-the-art GAN embedding method II2S~\citep{zhu2020improved}. A key insight is that we can use a heavily regularized version of the II2S GAN inversion method to do the reverse problem of transferring any related image (from a domain $B$) into the domain $A$, helping to characterize the semantic domain gap better than previous work.
Further extensions enable us to fine tune the modeling of the domain gap to explicitly model which attributes of the input image should be kept. Second, we propose multiple new regularizers to improve the quality. Third, we propose a technical improvement to the heuristic layer selection proposed in StyleGAN-NADA that is more straightforward and robust.


In summary, we make the following contributions:
\begin{enumerate}
  \item We reduce the mode collapse/overfitting problem which often occurs in one-shot and few-shot domain adaptation. Our results look similar to the target domain images with fewer artifacts. These results are also faithful to the identities of the source domain images and able to capture fine details.
  \item Our domain adaptation provides more freedom to control the “similarity” between images across domains that share a common latent-code, which makes a large number of downstream applications possible, e.g., pose adaptation, lighting adaptation, expression adaptation, texture adaptation, interpolation, and layer mixing, using state-of-the-art image editing frameworks.
\end{enumerate}

\section{Related Work}

\paragraph{Domain adaptation.}

Domain adaptation is the task of adapting a model to different domains. Different works in this area~\citep{bousmalis2016domain,Bousmalis_2017, na2020fixbi,Wang_2020,Kang_2019} try to learn diverse domain independent representations using the source domain to make predictions, such as image classification, in the target domains. More importantly, generating diverse representations of images by combining natural language supervision has been of interest to the computer vision and NLP research communities~\citep{frome2013devise}. Recently, OpenAI's Contrastive Language-Image Pretraining (CLIP)~\citep{radford2021learning} work established that transformer, and large datasets, could generate transferable visual models. In CLIP, both images and text are represented by high dimensional semantic-embedding vectors, which can then be used for zero-shot learning.

\paragraph{GAN-based domain adaptation.}
In the GAN domain, various models and training strategies have been proposed for few-shot domain adaptation tasks~\citep{Bousmalis_2017,NEURIPS2018_4e4e53aa,Li_2020,Liu_2019}. Most relevant to our work, the domain adaptation methods~\citep{patashnik2021styleclip,gal2021stylegannada,Jang_2021,Song:2021:AgileGAN} that build upon StyleGAN~\citep{karras2019style,Karras2019stylegan2,Karras2020ada} demonstrate impressive visual quality and semantic interpretability in the target domain. These methods can be broadly classified into few-shot and single-shot domain adaptation methods.

A notable few-shot method, StyleGAN-ADA~\citep{Karras2020ada} proposes an adaptive discriminator augmentation method to train StyleGAN on limited data. Another work, DiffAug~\citep{zhao2020differentiable}, applies differentiable transformations to the real and generated images for robust training. A discriminator related approach, FreezeD~\citep{mo2020freeze}, freezes lower layers of the discriminator to achieve domain adaptation. Toonify~\citep{gittoonify} interpolates between the model-weights of different generators to generate samples from a novel domain. A more recent work~\citep{ojha2021fewshot}, reduces overfitting on limited data by preserving the relative similarities and differences in the instances of samples in the source domain using cross domain correspondence.


\paragraph{Latent space interpretation and semantic editing.}

GAN interpretation and understanding of the latent space has been a topic of interest since the advent of GANs. Some notable works in this domain~\citep{bau2018gan,bau2019seeing,harkonen2020ganspace,shen2020interfacegan,tewari2020stylerig}  have led to many GAN-based image editing applications. More recent studies into the activation space of StyleGAN have demonstrated that the GAN can be exploited for downstream tasks like unsupervised and few-shot part segmentation~\citep{zhang2021datasetgan,tritrong2021repurposing,abdal2021labels4free, collins2020editing, bielski2019emergence}, extracting 3D models of the objects~\citep{pan2021do,chan2020pigan} and other semantic image editing applications~\citep{zhu2021barbershop, tan2020michigan, wu2020stylespace, patashnik2021styleclip}.

Image embedding is one of the approaches used to study the interpretability of the GANs. To enable the semantic editing of a given image using GANs, one needs to embed/project the image into its latent space. Image2StyleGAN~\citep{abdal2019image2stylegan} embeds images into the extended StyleGAN space called $W+$ space. Some followup works~\citep{zhu2020domain,richardson2020encoding,tewari2020pie} introduce regularizers and encoders to keep the latent code faithful to the original space of the StyleGAN. Improved-Image2StyleGAN (II2S)~\citep{zhu2020improved} uses $P_N$ space to regularize the embeddings for high-quality image reconstruction and image editing. We use this method to embed real images into the StyleGAN and show that our domain adaptation preserves the properties of the original StyleGAN in Sec~\ref{sec:results}.

Image editing is another tool to identify the concepts learned by a GAN. 
In the StyleGAN domain, recent works~\citep{harkonen2020ganspace,shen2020interfacegan,tewari2020stylerig,10.1145/3447648}  extract meaningful linear and non-linear paths in the latent space. InterfaceGAN~\citep{shen2020interfacegan}  finds linear directions to edit latent-codes in a supervised manner. On the other hand, GANSpace~\citep{harkonen2020ganspace} extracts unsupervised linear directions for editing using PCA in the $W$ space. Another framework, StyleRig~\citep{tewari2020stylerig}, maps the latent space of the GAN to a 3D model. StyleFlow~\citep{10.1145/3447648} extracts non-linear paths in the latent space to enable sequential image editing. In this work, we will use StyleFlow to test the semantic editing of our domain adapted images. 

In the area of text-based image editing, StyleCLIP~\citep{patashnik2021styleclip} extends CLIP to perform GAN-based image editing. StyleCLIP uses the CLIP embedding vector to traverse the StyleGAN manifold, by adjusting the latent-codes of a GAN, in order to make a generated image's CLIP embedding similar to the target vector, while remaining close to the input in latent space. A downside to this approach is that these edits are unable to shift the domain of a GAN outside its original manifold. However, their use of CLIP embeddings inspired StyleGAN-NADA~\citep{gal2021stylegannada}, which creates a new GAN using refinement learning to do zero-shot domain adaptation. Although unpublished, they also demonstrate one-shot domain adaptation in their accompanying code. The original and target domain are represented by CLIP text embeddings. The difference of the embeddings represents a direction used to shift the domains. Although in the accompanying source-code~\citep{gitnada}, they use a bootstrap-estimate of the mean CLIP image embedding of the original domain, and use a reference image or its CLIP image embedding to represent the new domain.

\section{Method}

\begin{figure}[htbp]
    \centering
    \includegraphics[width=0.85\columnwidth]{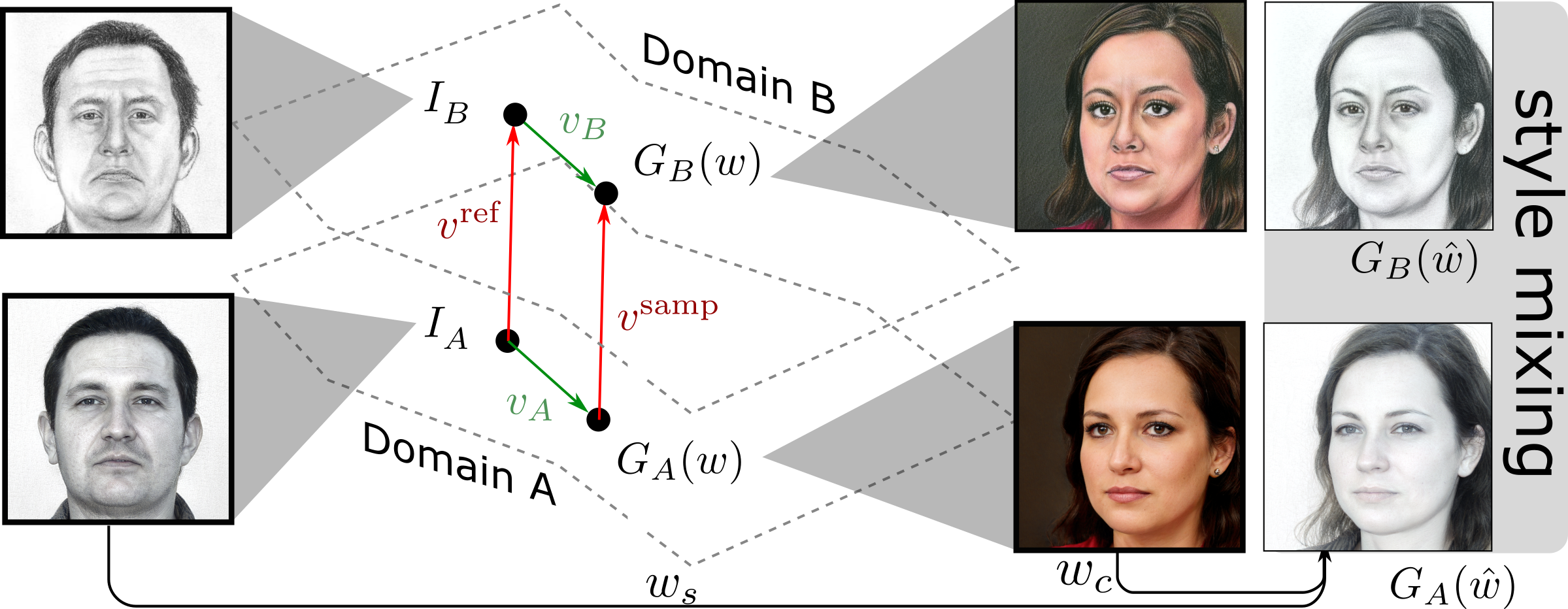}
    \caption{The vectors in the CLIP image embedding space, $E_I$, which control domain adaptation. Each domain is depicted here as a dashed outline; the vectors $v^\text{ref}$ and $v^\text{samp}$ cross between the two domains and are used to refine a generator for domain $B$. Corresponding images should be shifted in the same direction. The vectors $v_A$ and $v_B$ model important semantic differences within each domain that should also be preserved by domain transfer. $G_A(w)$ and $G_B(w)$ are corresponding images for an arbitrary latent-code $w$ encountered during training. Style mixing (shown on the right) shifts a part of the latent code towards the reference image effecting the result in both domains.}
    \label{fig:domain-xfer-clip-vectors}
\end{figure}

Our approach involves fine-tuning a GAN trained for some original domain $A$, e.g. FFHQ faces, to adapt it to a new related domain $B$. In our approach, the images in $A$ and the images in $B$ are related to each-other by a common latent code. Any image which can be generated or embedded in domain $A$ can be transferred to a corresponding and similar image in $B$. We use the CLIP embeddings as a semantic-space in order to model the difference between domains $A$ and $B$, and we use StyleGAN~\citep{STYLEGAN2018, Karras2019stylegan2} as the image generator.
A key to our approach is to preserve directions within  and across domains as illustrated in Fig.~\ref{fig:domain-xfer-clip-vectors}.
Before fine-tuning the GAN for domain $A$ (to obtain the GAN for domain $B$), we determine a domain-gap direction. This direction, called \RefAccrossDomainShift{}, is a vector in CLIP embedding space which points towards a reference image \RefImageB{} which is in domain $B$ from its corresponding image \RefImageA{} in which is in domain $A$.
We use the CLIP image-embedding model \CLIPEmbedImage{} to find
\begin{align}
    \RefAccrossDomainShift{} &= \CLIPEmbedImage{}(\RefImageB{}) - \CLIPEmbedImage{}(\RefImageA{}).  
\end{align}

Finding \RefImageA{} in domain A for a given image in domain B is a significant limitation in the current state of the art, StyleGAN-NADA~\citep{gal2021stylegannada}, as they use the mean of domain $A$. The mean of domain A is a very crude approximation for \RefImageA{}. Instead, we propose an inverse domain adaption step, by projecting the image \RefImageB{} into the domain $A$ to find a sample that is more similar and specific to the reference image than the mean of domain $A$. In principle, this problem is also a domain adaption problem similar to the problem we are trying to solve, just in the inverse direction. The major difference is that we have a pre-trained GAN available in domain A.

\begin{figure}[htbp]
    \centering
    \includegraphics[width=0.85\textwidth]{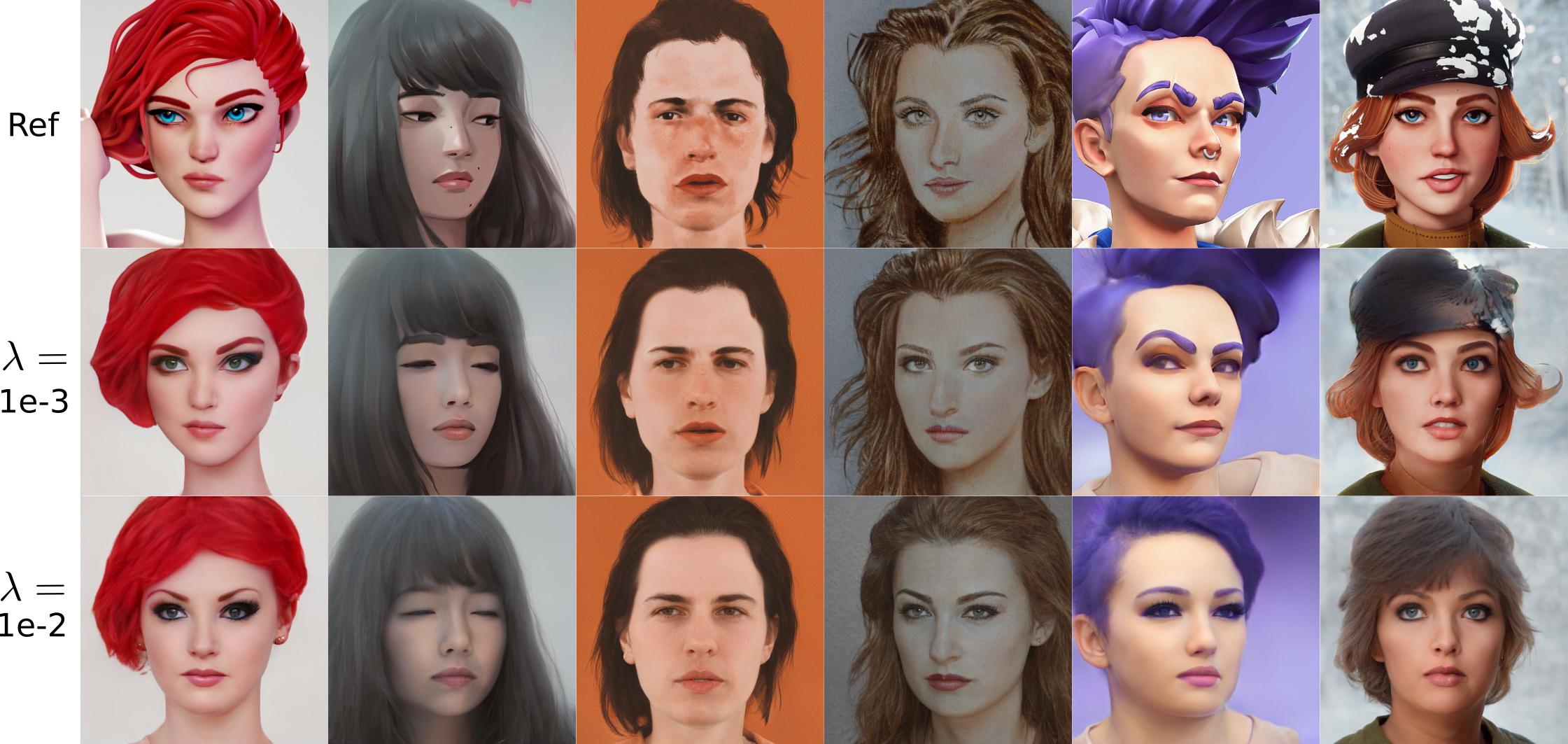}
    \caption{An illustration showing how II2S embeds \RefImageB{} in the original StyleGAN domain $A$, shown for two different values of $\lambda$. Reference images from other domains are shown in the top row. The value recommended by~\citet{zhu2020improved} is shown in the second row, and the value used in this work is shown in the third row. Although there is some subjectivity involved, we believe that the large value $\lambda=1\mathrm{e}{-2}$ is needed for II2S to find images that plausibly could belong to the domain $A$, which in this case is FFHQ faces. }
    \label{fig:ii2s-lambda}
\end{figure}

We use the II2S GAN-inversion method~\citep{zhu2020improved} in order to find a latent code for an image similar to \RefImageB{} that is plausibly within domain $A$. The I2S and II2S methods use  an extended version of $W$ space from StyleGAN2. The $W$ code is used 18 times, once for each style block in StyleGAN2. When allowing each element to vary independently, the resulting latent space is called $W+$ space~\cite{abdal2019image2stylegan, abdal2020image2stylegan++, zhu2020improved}. I2S showed that the additional degrees of freedom allow GAN inversion for a wider set of images with very detailed reconstruction capabilities, and II2S showed that an additional regularization term to keep the latent codes close to their original distribution made latent-code manipulation more robust. 
II2S uses a hyperparameter, $\lambda$, which can be increased in order to generate latent codes using more regularization, and therefore in higher density regions of the $W+$ latent space. The effect of this parameter is shown in Fig.~\ref{fig:ii2s-lambda}. The value suggested in the original work was $\lambda=0.001$, however, low values of lambda allow II2S to find latent codes that are too far away from the latent-codes produced by the mapping network of the original GAN and thus produce images that are less plausible to have come from domain $A$, underestimating the gap between domains. In the context of domain shift we find it is useful to use $\lambda=0.01$ as illustrated in Fig.~\ref{fig:ii2s-lambda}.  
The result is a latent code \RefWplus{} in $W+$ space which is shifted towards a high-density portion of the domain $A$.  
Then the image generated from that code, \RefImageA{}, is an image in domain $A$ that corresponds to \RefImageB{}.

\begin{figure}[htbp]
    \centering
    \includegraphics[width=0.95\columnwidth]{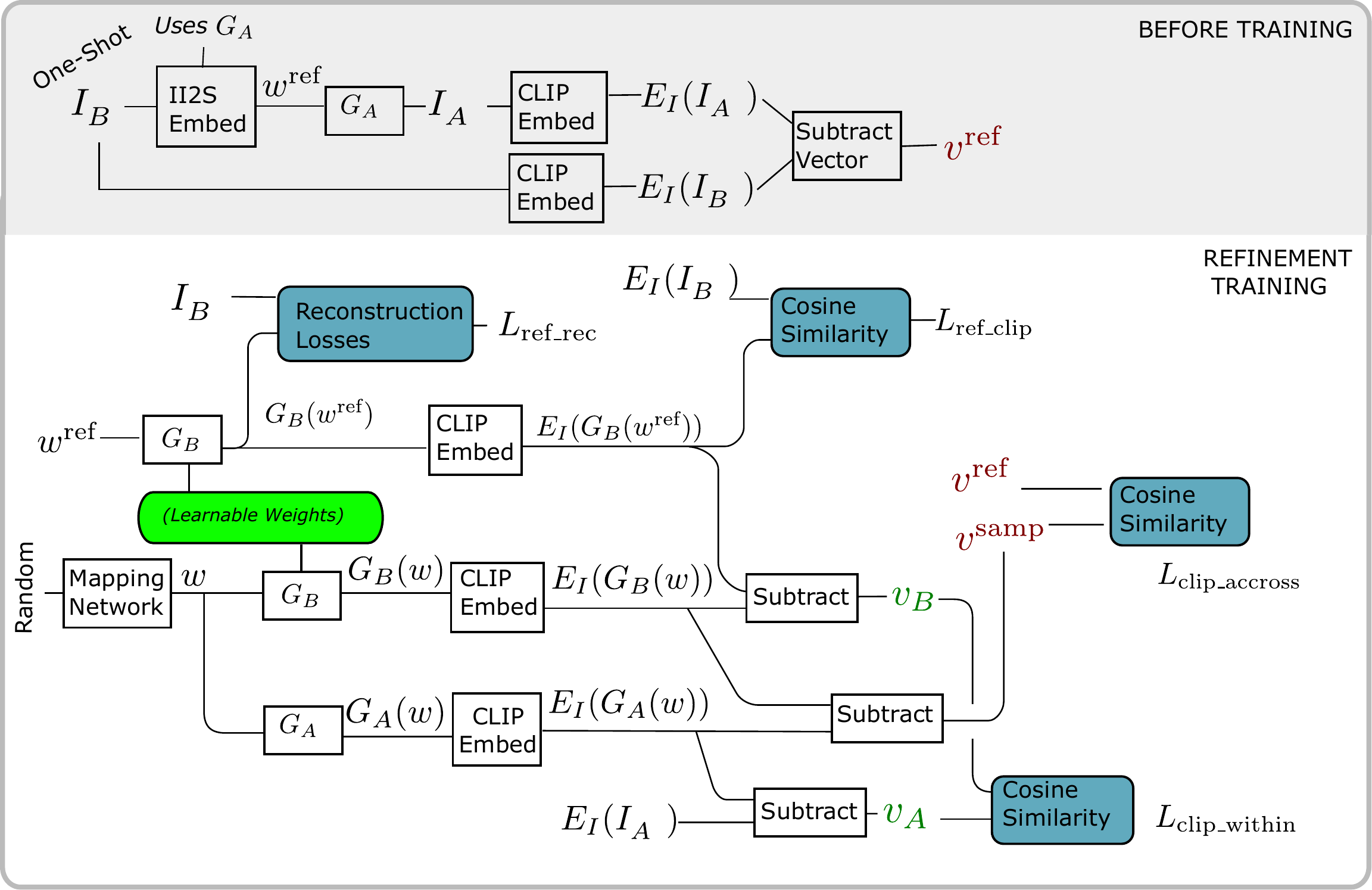}
    \caption{A process diagram for domain transfer. White rectangles indicate calculations, computed values are shown on the connecting lines. The four loss-calculations are indicated by blue rectangles, and the learnable weights of StyleGAN2 (all weights except the mapping network and the ToRGB layers) are indicated in green.  }
    \label{fig:training-process}
\end{figure}

\paragraph{Training}
As illustrated in Fig.~\ref{fig:ii2s-lambda}, we use II2S to find an image \RefImageA{} which we consider to be similar to \RefImageB{} but still plausibly within a domain $A$. In principle, it is possible that II2S finds \RefImageA{} so that \RefImageB{} is similar enough to be considered the same, in which case the two domains overlap. However, we are concerned with the cases where the domains are different, and the vector \RefAccrossDomainShift{} indicates the direction of a gap, or shift, between domain $A$ and domain $B$. We use refinement learning to train a new generator, $G_B$, so that images generated from $G_B$ are shifted parallel to \RefAccrossDomainShift{} in CLIP space, relative to images from $G_A$. The desired shift is indicated by the red arrows in Fig.~\ref{fig:domain-xfer-clip-vectors}.
During training, latent codes $w$ are generated using the mapping network of StyleGAN2.
Both $G_A$ and $G_B$  are used to generate images from the same latent code, but the weights of $G_A$ are frozen and only $G_B$ is updated during training. The goal of refinement learning is that $G_B$ will preserve semantic information that is \textit{within} domain $A$ but also that it will generate image shifted \textit{across} a gap between domains.
When refining the generator for domain $B$, we freeze the weights of the StyleGAN2 `ToRGB' layers, and the mapping network is also frozen. The overall process of training is illustrated in Fig.~\ref{fig:training-process}.

The goal of training is to shift CLIP embeddings from domain $A$ in a direction parallel to \RefAccrossDomainShift{}. We use the vector \SampleAccrossDomainShift{} to represent the current domain shift of the network $G_B$ during training, on a single sample. We have
\begin{align}
    \SampleAccrossDomainShift{}&=\CLIPEmbedImage{}(G_B(w))-\CLIPEmbedImage{}(G_A(w))
\end{align} 
as a cross-domain vector for corresponding images generated from the same $w$ latent code using the two generators. 
We use the loss
\begin{align}
    L_\text{clip\_across} &= 1-\text{sim}(\RefAccrossDomainShift{}, \SampleAccrossDomainShift{}),
\end{align}
where  $\text{sim}(\vb{a}, \vb{b}) = \frac{\vb{a}^T\vb{b}}{\|\vb{a}\|\|\vb{b}\|}$
is the cosine similarity score.  This loss term is minimized when the domain shift vectors are parallel.

It is important that the reference image \RefImageB{} matches the generated image, $G_B(\RefWplus{})$, both in a semantic sense, as measured by the similarity of the CLIP embeddings, and also in a visual sense. We accomplish this using two losses:  $L_\text{ref\_clip}$ and $L_\text{ref\_rec}$. The first loss measures the change in the CLIP-embeddings of the original and reconstructed reference image, 
\begin{align}
    L_\text{ref\_clip} &= 1-\text{sim}\left(
            \CLIPEmbedImage{}\left(\RefImageB{}\right), \CLIPEmbedImage{}\left(G_B(\RefWplus{})\right) 
        \right),
\end{align}
ensuring that the $G_B$ can reconstruct the embedding. Unlike $L_\text{clip\_accross}$, this loss term is not based on a \textit{change} in embeddings between the two domains, instead it guides $G_B$ by aligning it to a global embedding in CLIP space, ensuring that $I_B$ remains fixed in the domain of $G_B$.  

The second loss term is a reconstruction loss based on perceptual and pixel-level accuracy, 
\begin{align}
    L_\text{ref\_rec} &= 
        L_\text{PIPS}\left(\RefImageB{}, G_B(\RefWplus{})\right)
        + L_2\left(\RefImageB{}, G_B(\RefWplus{})\right) 
\end{align}
where
 $L_\text{PIPS}$ is the perceptual loss from~\citet{zhang2018unreasonable}, and $L_2$ is the squared euclidean difference between pixels.
 The purpose of this loss is to ensure that the visual, and not just the 
 semantic, qualities of the image are preserved. This is necessary in addition 
 to $ L_\text{ref\_clip}$ because, while the CLIP embeddings do capture many 
 semantic and visual qualities of the image, there are still many perceptually distinct 
 images that could produce the same CLIP embedding. This is visible in 
 Fig.~\ref{fig:Ablation_study}, without the reconstruction loss $G_B$ fails to 
 preserve some important visual qualities (such as symmetry) of the input. 
 
There is a tendency for GANs to reduce the variation during training, especially in few-shot fine-tuning.  We combat this by preserving the semantic information that is \textit{not} related to the domain gap.  A semantic change that is not related to the change in domains should not be affected by $G_B$. Therefore, the vector connecting the reference and sample images within the domain $A$ should be parallel to the corresponding vector in domain $B$.  Let
$ v_A = \CLIPEmbedImage{}(G_A(w)) - \CLIPEmbedImage{}(\RefImageA{}) $
be a vector connecting a sample image with latent-code $w$ to the reference image in the CLIP space.  This vector represents semantic changes that are \textit{within} domain $A$, and we want the matching semantic changes to occur within the domain $B$. Let 
$v_B = \CLIPEmbedImage{}(G_B(w)) - \CLIPEmbedImage{}(\RefImageB{})$
denote the corresponding vector in domain $B$.  We introduce the loss
\begin{align}
    L_\text{clip\_within}& = 1-\text{sim}(v_A, v_B),
\end{align}
which is minimized when the two within-domain changes are parallel. 

The final loss is then a weighted sum of losses
\begin{align}
L &= L_\text{clip\_across} + \lambda_\text{clip\_within}L_\text{clip\_within} + \lambda_\text{ref\_clip}L_\text{ref\_clip} + \lambda_\text{ref\_rec}L_\text{ref\_rec},    
\end{align}
with empirically determined weights of $\lambda_\text{clip\_within}=0.5$, $\lambda_\text{ref\_clip}=30$, and $\lambda_\text{ref\_rec}=10$. 
Together, these four loss terms guide the refinement process for $G_B$.  Among these losses, $L_\text{clip\_across}$ was proposed by StyleGAN-NADA~\citep{gal2021stylegannada}. The other losses are novel contributions of this work. 

\paragraph{Style Mixing}  
After the training step, the generator $G_B$ generates images that are semantically similar to the reference image \RefImageB{}. However, we have observed that the visual style may not be sufficiently similar. We attribute this to the idea that the target domain may be a \textit{subset} of the images produced by the new generator $G_B$.  This issue was addressed in StyleGAN-NADA~\citep{gal2021stylegannada} using a second latent-mining network in order to identify a distribution of latent codes within the domain of $G_B$ that better match the reference image. Our approach exploits the structure of latent codes in $W+$ space. Latent vectors in $W+$ space can be divided into 18 blocks of 512 elements, each impacting a different layer of StyleGAN2. Empirically, the latter blocks of the $W+$ code have been shown to have more effect on the style (e.g. texture and color) of the image whereas the earlier layers impact the coarse-structure or content~\citep{zhu2021barbershop} of the image.   We partition the latent code in the image into $w= (w_C,w_S)$ where $w_C$ consists of the first $m$ blocks of the $W+$ latent code that capture the content of the image, and $w_S$ consists of the remaining blocks and captures the style. In this work, we will use $m=7$ unless otherwise specified.  Then we transfer the style from a reference image using linear interpolation, to form $\hat{w} = (w_C, \hat{w}_S)$ where
\begin{align}
 \hat{w}_S &= (1-\alpha) w_S + \alpha(w_S^\text{ref}),
 \label{eg:eight}
\end{align}
and $w_S^\text{ref}$ is last $(18-m)$ blocks of $w^\text{ref}$.
Consider the distribution of images generated from random $w$ drawn according to the distribution of latent codes from the mapping network of StyleGAN2. 
If $\alpha=0$, then the distribution of images $G_B(\hat{w})$ includes the reference image, but encompasses a wide variety of other fine visual styles. If $\alpha=1$, then the images $G_B(\hat{w})$ will still have a diverse content, but they will all very closely follow the visual style of \RefImageB{}. An important application of this method is in conditional editing of real photographs. To achieve that, first we take a real input image $I_\text{real}$ and invert  it in domain $A$ using II2S on the generator $G_A$ in order to find a $W+$ latent code $w_\text{real}$. Then $G_B(w_\text{real})$ generates a corresponding image in domain $B$. We can then compute $\hat{w}_\text{real}$ by interpolating the style codes (\ref{eg:eight}) so that the final image $G_B(\hat{w}_\text{real})$ is similar to $I_\text{real}$ but has both content and the visual style shifted towards domain $B$.

\section{Results}
\label{sec:results}

In this section, we will show qualitative and quantitative results of our work. The only other published method that accomplishes similar one-shot GAN domain adaptation which we are aware of is~\citet{ojha2021fewshot}. They focus on few-shot domain adaptation, but they also demonstrate a capability to solve the one-shot problem. The most closely related work to our approach is StyleGAN-NADA~\citep{gal2021stylegannada}, which is unpublished at the time of submission, however we compare to it as the main competitor. 
The paper mainly discusses zero-shot domain adaptation, but the approach can also accomplish one-shot domain adaptation, as demonstrated in their accompanying source-code. Moreover, it demonstrates impressive improvements over the state of the art and even beats many SOTA few-shot methods considering the visual quality.
As our method can still significantly improve upon the results shown in StyleGAN-NADA, this underlines the importance of our idea in reducing overfitting.
We compare against additional approaches in the appendix. 

\paragraph{Training and Inference Time.}
Given a reference image, the training time for our method is about 15 minutes for 600 iterations on a single Titan~XP GPU using ADAM as an optimizer with the same settings as~\citet{gal2021stylegannada}. We use a batch size of 4. At inference time, there are different applications. In a basic operation, GAN generated images can be transferred with a single forward pass through a GAN generator network, which works in 0.34 seconds. Considering a more advanced operation, where existing photographs are embedded into a GAN latent space, the additional embedding time has to be considered. This embedding time is only 0.22 seconds using e4e~\citep{tov2021designing} and about two minutes using II2S~\citep{zhu2020improved}.

\begin{figure*}[th]
    \centering
    \includegraphics[width=\linewidth]{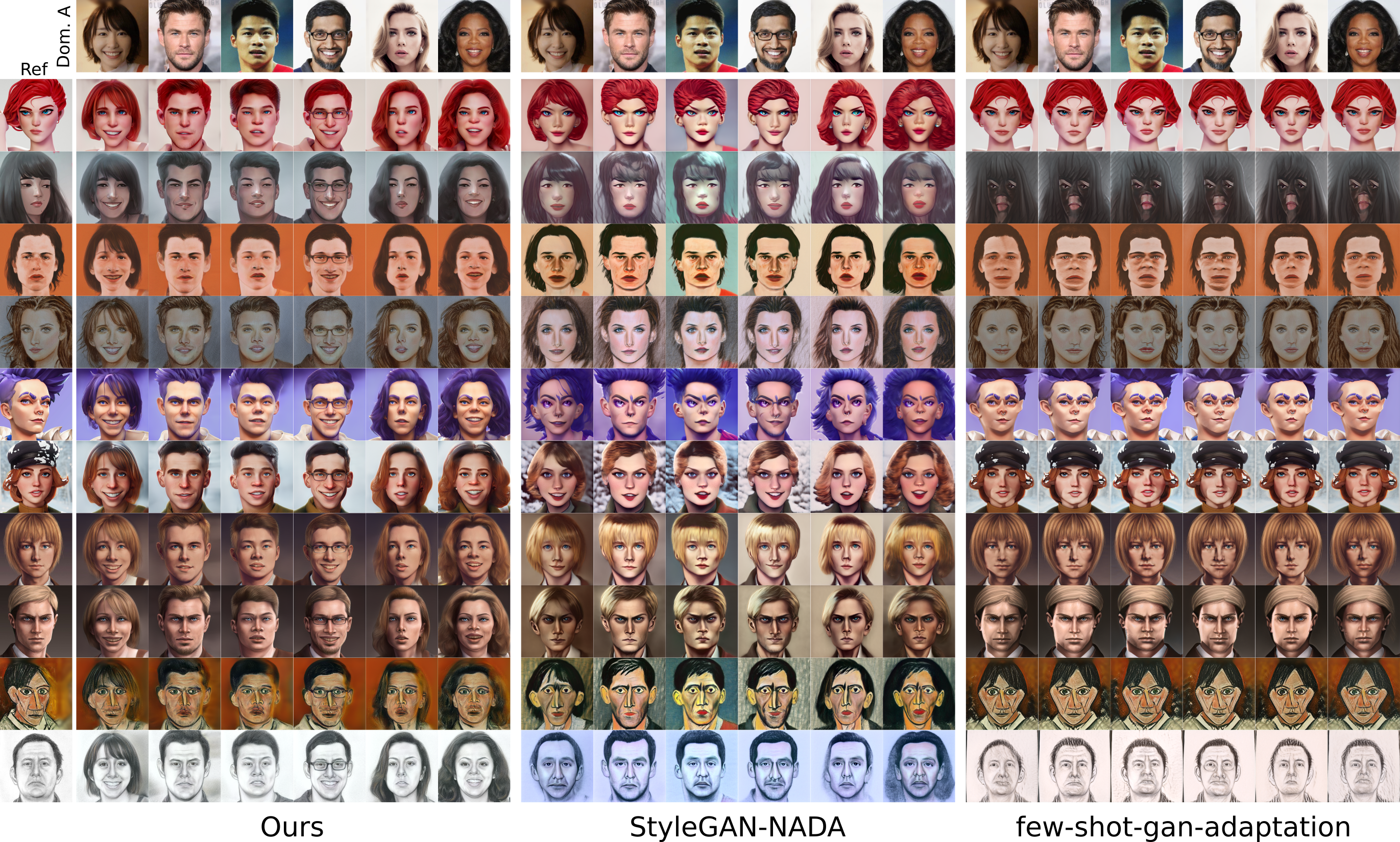}
    \caption{Comparison of our framework with state-of-the-art frameworks for StyleGAN domain adaptation. We compare with StyleGAN-NADA~\citep{gal2021stylegannada} and the few-shot method of~\citet{ojha2021fewshot}. Each row corresponds to a different reference image \RefImageB{}, and each column is a different real image $I_\text{real}$ from domain $A$. Notice that our method is able to match the styles of the reference images, while StyleGAN-NADA fails to maintain the content of the images in domain $A$ (for example the identity of a person is lost). On the other hand, the few-shot method suffers from severe mode collapse.}
    \label{fig:Comparison_with_sota}
\end{figure*}

\paragraph{Visual Evaluation.}
In Fig.~\ref{fig:Comparison_with_sota}, we show a comparison of our results on faces against the two most relevant competing methods -- StyleGAN-NADA~\citep{gal2021stylegannada} and few-shot-domain-adaptation~\citep{ojha2021fewshot}. The results show that our method remains faithful to the original identity of the embedded images in domain $A$, while the other two methods suffer from overfitting, i.e., collapsing to narrow distributions which do not preserve salient features (for example the identity of a person).
We show additional visual results in the supplemental materials, including results on cars and dogs and results for fine-tuning the domain adaptation.

\paragraph{User Study.}
We also perform a user study by collecting 187 responses from Amazon Mechanical Turk in order to compare the visual quality and the domain transfer capabilities of our framework compared to the competing methods. 
When asked which method generates higher quality images from domain $B$, 73\% of users preferred our approach to StyleGAN-NADA, and 77\% selected ours over Few-shot~\citep{ojha2021fewshot}. 
When asked which method is better at maintaining the similarity to a corresponding source image in domain $A$, we found that 80\% of the responses chose our approach over StyleGAN-NADA, and 91\% preferred our approach to Few-shot. 
Our method outperforms the competing works in terms of the quality of the generated image, and the similarity of the generated image to the source image from domain $A$.  According to the user study, the other methods produced images that are more similar to \RefImageB{}, but that is also an indication of overfitting and mode collapse. 

 

 



\begin{figure*}[htb]
    \centering
    \includegraphics[width=\linewidth]{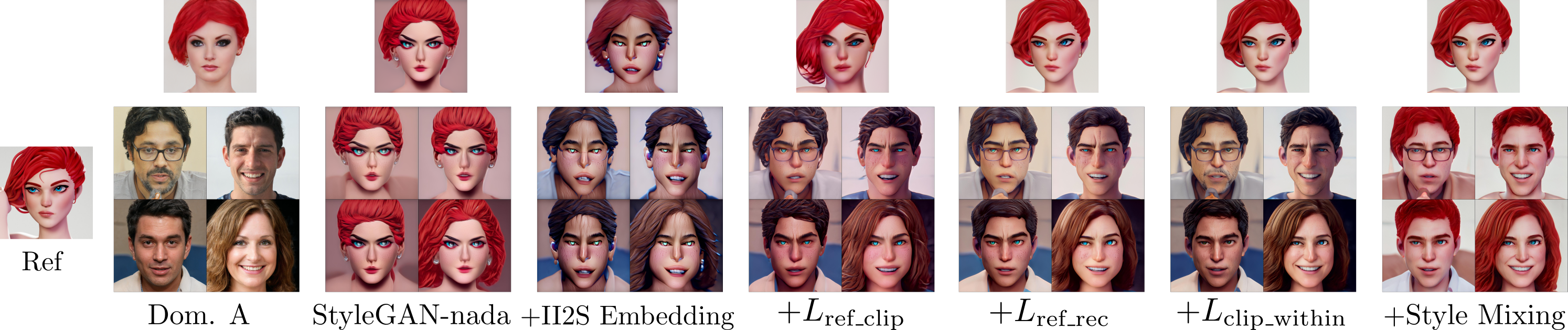}
    \caption{Ablation study of the losses and style mixing used in our framework. From left to right: the reference image \RefImageA{} and several images from domain $A$, the baseline approach (StyleGAN-NADA), adding II2S instead of using the mean of domain $A$, adding $L_\text{ref\_clip}$, $L_\text{clip\_within}$, and then using style mixing. The top row shows reconstructions of the image \RefImageA{} using $G_B$. }
    \label{fig:Ablation_study}
\end{figure*}

\paragraph{Ablation study.}
We perform an ablation study to evaluate each component of our framework. In Fig.~\ref{fig:Ablation_study}, we show the effect of II2S embedding, different losses and style mixing/interpolation on the output. 

\begin{figure*}[th]
    \centering
    \includegraphics[width=\linewidth]{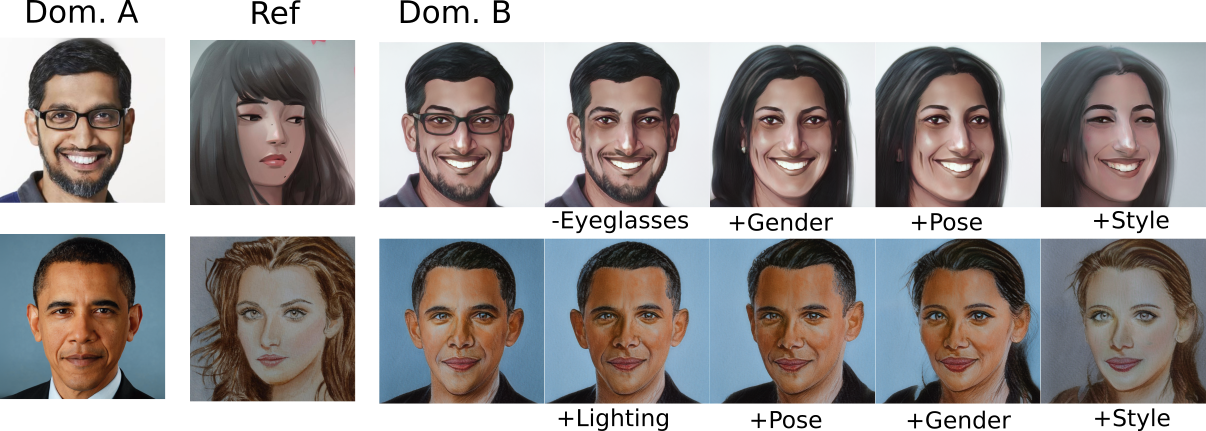}
    \caption{Image editing capabilities of the new domain $B$ using StyleFlow~\citep{10.1145/3447648}. 
    This figure shows the editing results of the embedded real image $I_{real}$ transferred to domain $B$. Notice that our method preserves the semantic properties of the original StyleGAN.
    }

    \label{fig:Attributes_control}
\end{figure*}

\paragraph{Image editing capabilities.}
Another important aspect of our method is that we are able to preserve the semantic properties of the original StyleGAN (domain $A$) in domain $B$. We can make edits to the images in domain $B$ via the learned generator $G_B$ without retraining the image editing frameworks on the new domain. Fig.~\ref{fig:Attributes_control} shows image editing capabilities in the new domain $B$.
We use StyleFlow edits such as lighting, pose, gender etc. to show the fine-grained edits possible in the new domain.

\begin{figure*}[th]
    \centering
    \includegraphics[width=\linewidth]{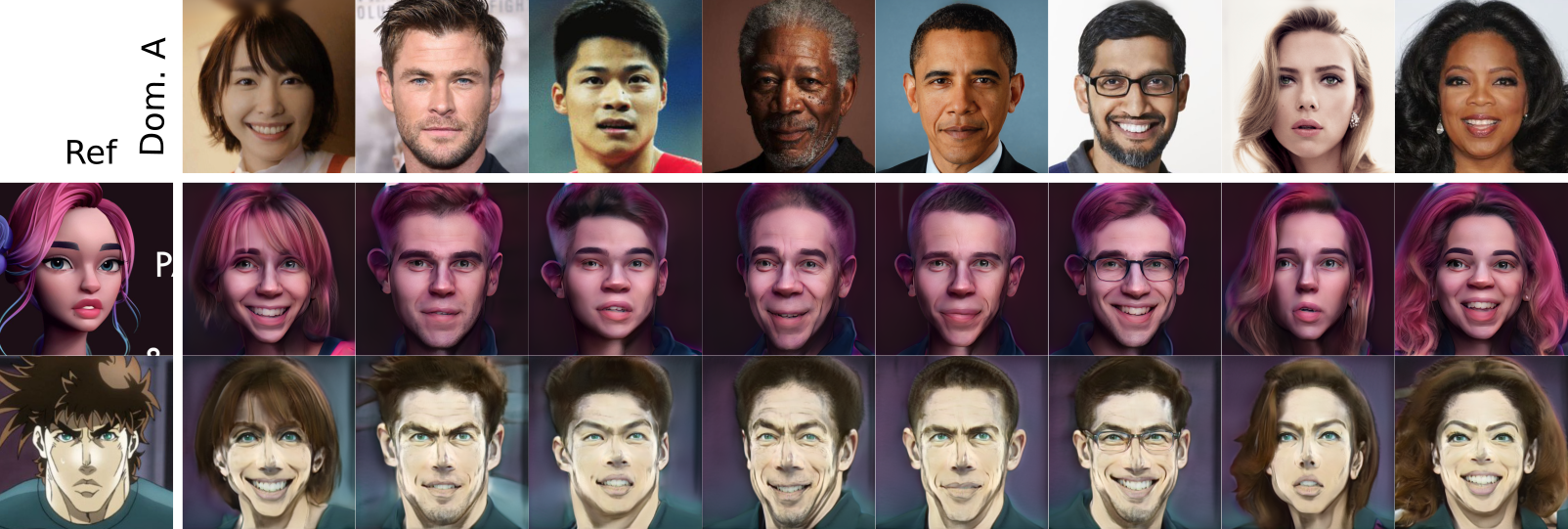}
    \caption{Some failure cases of our method. In these examples, we observe that the identity of the face is compromised a bit more than in typical examples of our method.}
    \label{fig:Failded_case}
\end{figure*}

\paragraph{Limitations}
Our method has several limitations (See Fig.~\ref{fig:Failded_case}). Some of these limitations are inherent due to the challenging nature of the problem of single-shot domain adaptation. Other limitations can be addressed in future work. 
First, when we find the initial image in domain $A$ that corresponds to the input in domain $B$, we do not attempt to control for the semantic similarity. Future work should encourage the images to have similar semantics. 
Second, we can only transfer between related domains. For example transferring FFHQ faces into the domain of cars is not explored in this paper. 
Third, also relevant to the original distribution of the StyleGAN, embeddings into the StyleGAN work best when the objects are transformed to the canonical positions (for example face poses that are the same as FFHQ). Extreme poses of the objects in the reference images sometimes fail.

\section{Conclusions}
We propose a novel method for single shot domain adaption. The main achievement of this work is to obtain results of unprecedented quality while reducing overfitting observed in previous work.
The technical key components of our work are a method to model the domain gap as vector in CLIP embedding space, a way to preserve within-domain variation, and  several extensions for fine-grained attribute-based control. We also introduce several new regularizers and a style mixing approach.

\newpage
\bibliography{iclr2022_conference}

\begin{thebibliography}{51}
\providecommand{\natexlab}[1]{#1}
\providecommand{\url}[1]{\texttt{#1}}
\expandafter\ifx\csname urlstyle\endcsname\relax
  \providecommand{\doi}[1]{doi: #1}\else
  \providecommand{\doi}{doi: \begingroup \urlstyle{rm}\Url}\fi

\bibitem[Abdal et~al.(2019)Abdal, Qin, and Wonka]{abdal2019image2stylegan}
Rameen Abdal, Yipeng Qin, and Peter Wonka.
\newblock Image2stylegan: How to embed images into the stylegan latent space?
\newblock In \emph{Proceedings of the IEEE/CVF International Conference on
  Computer Vision}, pp.\  4432--4441, 2019.

\bibitem[Abdal et~al.(2020)Abdal, Qin, and Wonka]{abdal2020image2stylegan++}
Rameen Abdal, Yipeng Qin, and Peter Wonka.
\newblock Image2stylegan++: How to edit the embedded images?
\newblock In \emph{Proceedings of the IEEE/CVF Conference on Computer Vision
  and Pattern Recognition}, pp.\  8296--8305, 2020.

\bibitem[Abdal et~al.(2021{\natexlab{a}})Abdal, Zhu, Mitra, and
  Wonka]{abdal2021labels4free}
Rameen Abdal, Peihao Zhu, Niloy Mitra, and Peter Wonka.
\newblock Labels4free: Unsupervised segmentation using stylegan,
  2021{\natexlab{a}}.

\bibitem[Abdal et~al.(2021{\natexlab{b}})Abdal, Zhu, Mitra, and
  Wonka]{10.1145/3447648}
Rameen Abdal, Peihao Zhu, Niloy~J. Mitra, and Peter Wonka.
\newblock Styleflow: Attribute-conditioned exploration of stylegan-generated
  images using conditional continuous normalizing flows.
\newblock \emph{ACM Trans. Graph.}, 40\penalty0 (3), May 2021{\natexlab{b}}.
\newblock ISSN 0730-0301.
\newblock \doi{10.1145/3447648}.
\newblock URL \url{https://doi.org/10.1145/3447648}.

\bibitem[Bau et~al.(2018)Bau, Zhu, Strobelt, Zhou, Tenenbaum, Freeman, and
  Torralba]{bau2018gan}
David Bau, Jun-Yan Zhu, Hendrik Strobelt, Bolei Zhou, Joshua~B. Tenenbaum,
  William~T. Freeman, and Antonio Torralba.
\newblock Gan dissection: Visualizing and understanding generative adversarial
  networks, 2018.

\bibitem[Bau et~al.(2019)Bau, Zhu, Wulff, Peebles, Strobelt, Zhou, and
  Torralba]{bau2019seeing}
David Bau, Jun-Yan Zhu, Jonas Wulff, William Peebles, Hendrik Strobelt, Bolei
  Zhou, and Antonio Torralba.
\newblock Seeing what a gan cannot generate.
\newblock In \emph{Proceedings of the IEEE/CVF International Conference on
  Computer Vision}, pp.\  4502--4511, 2019.

\bibitem[Bielski \& Favaro(2019)Bielski and Favaro]{bielski2019emergence}
Adam Bielski and Paolo Favaro.
\newblock Emergence of object segmentation in perturbed generative models.
\newblock \emph{arXiv preprint arXiv:1905.12663}, 2019.

\bibitem[Bousmalis et~al.(2016)Bousmalis, Trigeorgis, Silberman, Krishnan, and
  Erhan]{bousmalis2016domain}
Konstantinos Bousmalis, George Trigeorgis, Nathan Silberman, Dilip Krishnan,
  and Dumitru Erhan.
\newblock Domain separation networks, 2016.

\bibitem[Bousmalis et~al.(2017)Bousmalis, Silberman, Dohan, Erhan, and
  Krishnan]{Bousmalis_2017}
Konstantinos Bousmalis, Nathan Silberman, David Dohan, Dumitru Erhan, and Dilip
  Krishnan.
\newblock Unsupervised pixel-level domain adaptation with generative
  adversarial networks.
\newblock \emph{2017 IEEE Conference on Computer Vision and Pattern Recognition
  (CVPR)}, Jul 2017.
\newblock \doi{10.1109/cvpr.2017.18}.
\newblock URL \url{http://dx.doi.org/10.1109/CVPR.2017.18}.

\bibitem[Chan et~al.(2020)Chan, Monteiro, Kellnhofer, Wu, and
  Wetzstein]{chan2020pigan}
Eric~R. Chan, Marco Monteiro, Petr Kellnhofer, Jiajun Wu, and Gordon Wetzstein.
\newblock pi-gan: Periodic implicit generative adversarial networks for
  3d-aware image synthesis, 2020.

\bibitem[Chefer et~al.(2021)Chefer, Benaim, Paiss, and
  Wolf]{chefer2021targetclip}
Hila Chefer, Sagie Benaim, Roni Paiss, and Lior Wolf.
\newblock Image-based clip-guided essence transfer.
\newblock \emph{arXiv preprint arXiv: 2110.12427}, 2021.

\bibitem[Collins et~al.(2020)Collins, Bala, Price, and
  Süsstrunk]{collins2020editing}
Edo Collins, Raja Bala, Bob Price, and Sabine Süsstrunk.
\newblock Editing in style: Uncovering the local semantics of gans, 2020.

\bibitem[Frome et~al.(2013)Frome, Corrado, Shlens, Bengio, Dean, Ranzato, and
  Mikolov]{frome2013devise}
Andrea Frome, Greg Corrado, Jonathon Shlens, Samy Bengio, Jeffrey Dean,
  Marc’Aurelio Ranzato, and Tomas Mikolov.
\newblock Devise: A deep visual-semantic embedding model.
\newblock 2013.

\bibitem[Gal et~al.(2021)Gal, Patashnik, Maron, Chechik, and
  Cohen-Or]{gal2021stylegannada}
Rinon Gal, Or~Patashnik, Haggai Maron, Gal Chechik, and Daniel Cohen-Or.
\newblock Stylegan-nada: Clip-guided domain adaptation of image generators,
  2021.

\bibitem[Gatys et~al.(2016)Gatys, Ecker, and Bethge]{gatys2016image}
Leon~A Gatys, Alexander~S Ecker, and Matthias Bethge.
\newblock Image style transfer using convolutional neural networks.
\newblock In \emph{Proceedings of the IEEE conference on computer vision and
  pattern recognition}, pp.\  2414--2423, 2016.

\bibitem[H{\"a}rk{\"o}nen et~al.(2020)H{\"a}rk{\"o}nen, Hertzmann, Lehtinen,
  and Paris]{harkonen2020ganspace}
Erik H{\"a}rk{\"o}nen, Aaron Hertzmann, Jaakko Lehtinen, and Sylvain Paris.
\newblock Ganspace: Discovering interpretable gan controls.
\newblock \emph{arXiv preprint arXiv:2004.02546}, 2020.

\bibitem[Huang \& Belongie(2017)Huang and Belongie]{Adain2017}
Xun Huang and Serge Belongie.
\newblock Arbitrary style transfer in real-time with adaptive instance
  normalization.
\newblock In \emph{ICCV}, 2017.

\bibitem[Jang et~al.(2021)Jang, Ju, Jung, Yang, Tong, and Lee]{Jang_2021}
Wonjong Jang, Gwangjin Ju, Yucheol Jung, Jiaolong Yang, Xin Tong, and Seungyong
  Lee.
\newblock Stylecarigan.
\newblock \emph{ACM Transactions on Graphics}, 40\penalty0 (4):\penalty0
  1–16, Aug 2021.
\newblock ISSN 1557-7368.
\newblock \doi{10.1145/3450626.3459860}.
\newblock URL \url{http://dx.doi.org/10.1145/3450626.3459860}.

\bibitem[justinpinkney/toonify()]{gittoonify}
justinpinkney/toonify.
\newblock Toonify.
\newblock \url{https://github.com/justinpinkney/toonify}.

\bibitem[Kang et~al.(2019)Kang, Jiang, Yang, and Hauptmann]{Kang_2019}
Guoliang Kang, Lu~Jiang, Yi~Yang, and Alexander~G. Hauptmann.
\newblock Contrastive adaptation network for unsupervised domain adaptation.
\newblock \emph{2019 IEEE/CVF Conference on Computer Vision and Pattern
  Recognition (CVPR)}, Jun 2019.
\newblock \doi{10.1109/cvpr.2019.00503}.
\newblock URL \url{http://dx.doi.org/10.1109/CVPR.2019.00503}.

\bibitem[Karras et~al.(2018)Karras, Laine, and Aila]{STYLEGAN2018}
Tero Karras, Samuli Laine, and Timo Aila.
\newblock A style-based generator architecture for generative adversarial
  networks.
\newblock \emph{arXiv preprint arXiv:1812.04948}, 2018.

\bibitem[Karras et~al.(2019)Karras, Laine, and Aila]{karras2019style}
Tero Karras, Samuli Laine, and Timo Aila.
\newblock A style-based generator architecture for generative adversarial
  networks.
\newblock In \emph{Proceedings of the IEEE/CVF Conference on Computer Vision
  and Pattern Recognition}, pp.\  4401--4410, 2019.

\bibitem[Karras et~al.(2020{\natexlab{a}})Karras, Aittala, Hellsten, Laine,
  Lehtinen, and Aila]{Karras2020ada}
Tero Karras, Miika Aittala, Janne Hellsten, Samuli Laine, Jaakko Lehtinen, and
  Timo Aila.
\newblock Training generative adversarial networks with limited data.
\newblock In \emph{Proc. NeurIPS}, 2020{\natexlab{a}}.

\bibitem[Karras et~al.(2020{\natexlab{b}})Karras, Laine, Aittala, Hellsten,
  Lehtinen, and Aila]{Karras2019stylegan2}
Tero Karras, Samuli Laine, Miika Aittala, Janne Hellsten, Jaakko Lehtinen, and
  Timo Aila.
\newblock Analyzing and improving the image quality of {StyleGAN}.
\newblock In \emph{Proc. CVPR}, 2020{\natexlab{b}}.

\bibitem[Karras et~al.(2021)Karras, Aittala, Laine, H\"{a}rk\"{o}nen, Hellsten,
  Lehtinen, and Aila]{Karras2021}
Tero Karras, Miika Aittala, Samuli Laine, Erik H\"{a}rk\"{o}nen, Janne
  Hellsten, Jaakko Lehtinen, and Timo Aila.
\newblock Alias-free generative adversarial networks.
\newblock \emph{CoRR}, abs/2106.12423, 2021.

\bibitem[Li et~al.(2020)Li, Zhang, Li, and Fu]{Li_2020}
Kai Li, Yulun Zhang, Kunpeng Li, and Yun Fu.
\newblock Adversarial feature hallucination networks for few-shot learning.
\newblock \emph{2020 IEEE/CVF Conference on Computer Vision and Pattern
  Recognition (CVPR)}, Jun 2020.
\newblock \doi{10.1109/cvpr42600.2020.01348}.
\newblock URL \url{http://dx.doi.org/10.1109/cvpr42600.2020.01348}.

\bibitem[Liu et~al.(2019)Liu, Huang, Mallya, Karras, Aila, Lehtinen, and
  Kautz]{Liu_2019}
Ming-Yu Liu, Xun Huang, Arun Mallya, Tero Karras, Timo Aila, Jaakko Lehtinen,
  and Jan Kautz.
\newblock Few-shot unsupervised image-to-image translation.
\newblock \emph{2019 IEEE/CVF International Conference on Computer Vision
  (ICCV)}, Oct 2019.
\newblock \doi{10.1109/iccv.2019.01065}.
\newblock URL \url{http://dx.doi.org/10.1109/ICCV.2019.01065}.

\bibitem[Mo et~al.(2020)Mo, Cho, and Shin]{mo2020freeze}
Sangwoo Mo, Minsu Cho, and Jinwoo Shin.
\newblock Freeze the discriminator: a simple baseline for fine-tuning gans,
  2020.

\bibitem[Na et~al.(2020)Na, Jung, Chang, and Hwang]{na2020fixbi}
Jaemin Na, Heechul Jung, Hyung~Jin Chang, and Wonjun Hwang.
\newblock Fixbi: Bridging domain spaces for unsupervised domain adaptation,
  2020.

\bibitem[Ojha et~al.(2021)Ojha, Li, Lu, Efros, Lee, Shechtman, and
  Zhang]{ojha2021fewshot}
Utkarsh Ojha, Yijun Li, Jingwan Lu, Alexei~A. Efros, Yong~Jae Lee, Eli
  Shechtman, and Richard Zhang.
\newblock Few-shot image generation via cross-domain correspondence, 2021.

\bibitem[Pan et~al.(2021)Pan, Dai, Liu, Loy, and Luo]{pan2021do}
Xingang Pan, Bo~Dai, Ziwei Liu, Chen~Change Loy, and Ping Luo.
\newblock Do 2d {\{}gan{\}}s know 3d shape? unsupervised 3d shape
  reconstruction from 2d image {\{}gan{\}}s.
\newblock In \emph{International Conference on Learning Representations}, 2021.
\newblock URL \url{https://openreview.net/forum?id=FGqiDsBUKL0}.

\bibitem[Patashnik et~al.(2021)Patashnik, Wu, Shechtman, Cohen-Or, and
  Lischinski]{patashnik2021styleclip}
Or~Patashnik, Zongze Wu, Eli Shechtman, Daniel Cohen-Or, and Dani Lischinski.
\newblock Styleclip: Text-driven manipulation of stylegan imagery, 2021.

\bibitem[Radford et~al.(2021)Radford, Kim, Hallacy, Ramesh, Goh, Agarwal,
  Sastry, Askell, Mishkin, Clark, Krueger, and Sutskever]{radford2021learning}
Alec Radford, Jong~Wook Kim, Chris Hallacy, Aditya Ramesh, Gabriel Goh,
  Sandhini Agarwal, Girish Sastry, Amanda Askell, Pamela Mishkin, Jack Clark,
  Gretchen Krueger, and Ilya Sutskever.
\newblock Learning transferable visual models from natural language
  supervision, 2021.

\bibitem[Richardson et~al.(2020)Richardson, Alaluf, Patashnik, Nitzan, Azar,
  Shapiro, and Cohen-Or]{richardson2020encoding}
Elad Richardson, Yuval Alaluf, Or~Patashnik, Yotam Nitzan, Yaniv Azar, Stav
  Shapiro, and Daniel Cohen-Or.
\newblock Encoding in style: a stylegan encoder for image-to-image translation.
\newblock \emph{arXiv preprint arXiv:2008.00951}, 2020.

\bibitem[rinongal/StyleGAN NADA()]{gitnada}
rinongal/StyleGAN NADA.
\newblock Stylegan-nada.
\newblock \url{https://github.com/rinongal/StyleGAN-nada}.

\bibitem[Shen et~al.(2020)Shen, Yang, Tang, and Zhou]{shen2020interfacegan}
Yujun Shen, Ceyuan Yang, Xiaoou Tang, and Bolei Zhou.
\newblock Interfacegan: Interpreting the disentangled face representation
  learned by gans.
\newblock \emph{IEEE Transactions on Pattern Analysis and Machine
  Intelligence}, 2020.

\bibitem[Song et~al.(2021)Song, Luo, Liu, Ma, Lai, Zheng, and
  Cham]{Song:2021:AgileGAN}
Guoxian Song, Linjie Luo, Jing Liu, Wan-Chun Ma, Chunpong Lai, Chuanxia Zheng,
  and Tat-Jen Cham.
\newblock Agilegan: Stylizing portraits by inversion-consistent transfer
  learning.
\newblock \emph{ACM Transactions on Graphics (Proc. SIGGRAPH)}, jul 2021.

\bibitem[Tan et~al.(2020)Tan, Chai, Chen, Liao, Chu, Yuan, Tulyakov, and
  Yu]{tan2020michigan}
Zhentao Tan, Menglei Chai, Dongdong Chen, Jing Liao, Qi~Chu, Lu~Yuan, Sergey
  Tulyakov, and Nenghai Yu.
\newblock Michigan: Multi-input-conditioned hair image generation for portrait
  editing.
\newblock \emph{ACM Transactions on Graphics (TOG)}, 39\penalty0 (4):\penalty0
  1--13, 2020.

\bibitem[Tewari et~al.(2020{\natexlab{a}})Tewari, Elgharib, Bharaj, Bernard,
  Seidel, P{\'e}rez, Zollhofer, and Theobalt]{tewari2020stylerig}
Ayush Tewari, Mohamed Elgharib, Gaurav Bharaj, Florian Bernard, Hans-Peter
  Seidel, Patrick P{\'e}rez, Michael Zollhofer, and Christian Theobalt.
\newblock Stylerig: Rigging stylegan for 3d control over portrait images.
\newblock In \emph{Proceedings of the IEEE/CVF Conference on Computer Vision
  and Pattern Recognition}, pp.\  6142--6151, 2020{\natexlab{a}}.

\bibitem[Tewari et~al.(2020{\natexlab{b}})Tewari, Elgharib, BR, Bernard,
  Seidel, P{\'e}rez, Z{\"o}llhofer, and Theobalt]{tewari2020pie}
Ayush Tewari, Mohamed Elgharib, Mallikarjun BR, Florian Bernard, Hans-Peter
  Seidel, Patrick P{\'e}rez, Michael Z{\"o}llhofer, and Christian Theobalt.
\newblock Pie: Portrait image embedding for semantic control.
\newblock volume~39, December 2020{\natexlab{b}}.
\newblock \doi{10.1145/3414685.3417803}.

\bibitem[Tov et~al.(2021)Tov, Alaluf, Nitzan, Patashnik, and
  Cohen-Or]{tov2021designing}
Omer Tov, Yuval Alaluf, Yotam Nitzan, Or~Patashnik, and Daniel Cohen-Or.
\newblock Designing an encoder for stylegan image manipulation.
\newblock \emph{arXiv preprint arXiv:2102.02766}, 2021.

\bibitem[Tritrong et~al.(2021)Tritrong, Rewatbowornwong, and
  Suwajanakorn]{tritrong2021repurposing}
Nontawat Tritrong, Pitchaporn Rewatbowornwong, and Supasorn Suwajanakorn.
\newblock Repurposing gans for one-shot semantic part segmentation, 2021.

\bibitem[Wang \& Breckon(2020)Wang and Breckon]{Wang_2020}
Qian Wang and Toby Breckon.
\newblock Unsupervised domain adaptation via structured prediction based
  selective pseudo-labeling.
\newblock \emph{Proceedings of the AAAI Conference on Artificial Intelligence},
  34\penalty0 (04):\penalty0 6243–6250, Apr 2020.
\newblock ISSN 2159-5399.
\newblock \doi{10.1609/aaai.v34i04.6091}.
\newblock URL \url{http://dx.doi.org/10.1609/AAAI.V34I04.6091}.

\bibitem[Wu et~al.(2020)Wu, Lischinski, and Shechtman]{wu2020stylespace}
Zongze Wu, Dani Lischinski, and Eli Shechtman.
\newblock Stylespace analysis: Disentangled controls for stylegan image
  generation.
\newblock \emph{arXiv preprint arXiv:2011.12799}, 2020.

\bibitem[Zhang et~al.(2018)Zhang, Isola, Efros, Shechtman, and
  Wang]{zhang2018unreasonable}
Richard Zhang, Phillip Isola, Alexei~A. Efros, Eli Shechtman, and Oliver Wang.
\newblock The unreasonable effectiveness of deep features as a perceptual
  metric, 2018.

\bibitem[ZHANG et~al.(2018)ZHANG, Che, Ghahramani, Bengio, and
  Song]{NEURIPS2018_4e4e53aa}
Ruixiang ZHANG, Tong Che, Zoubin Ghahramani, Yoshua Bengio, and Yangqiu Song.
\newblock Metagan: An adversarial approach to few-shot learning.
\newblock In S.~Bengio, H.~Wallach, H.~Larochelle, K.~Grauman, N.~Cesa-Bianchi,
  and R.~Garnett (eds.), \emph{Advances in Neural Information Processing
  Systems}, volume~31. Curran Associates, Inc., 2018.
\newblock URL
  \url{https://proceedings.neurips.cc/paper/2018/file/4e4e53aa080247bc31d0eb4e7aeb07a0-Paper.pdf}.

\bibitem[Zhang et~al.(2021)Zhang, Ling, Gao, Yin, Lafleche, Barriuso, Torralba,
  and Fidler]{zhang2021datasetgan}
Yuxuan Zhang, Huan Ling, Jun Gao, Kangxue Yin, Jean-Francois Lafleche, Adela
  Barriuso, Antonio Torralba, and Sanja Fidler.
\newblock Datasetgan: Efficient labeled data factory with minimal human effort,
  2021.

\bibitem[Zhao et~al.(2020)Zhao, Liu, Lin, Zhu, and Han]{zhao2020differentiable}
Shengyu Zhao, Zhijian Liu, Ji~Lin, Jun-Yan Zhu, and Song Han.
\newblock Differentiable augmentation for data-efficient gan training, 2020.

\bibitem[Zhu et~al.(2020{\natexlab{a}})Zhu, Shen, Zhao, and
  Zhou]{zhu2020domain}
Jiapeng Zhu, Yujun Shen, Deli Zhao, and Bolei Zhou.
\newblock In-domain gan inversion for real image editing.
\newblock In \emph{European Conference on Computer Vision}, pp.\  592--608.
  Springer, 2020{\natexlab{a}}.

\bibitem[Zhu et~al.(2020{\natexlab{b}})Zhu, Abdal, Qin, Femiani, and
  Wonka]{zhu2020improved}
Peihao Zhu, Rameen Abdal, Yipeng Qin, John Femiani, and Peter Wonka.
\newblock Improved stylegan embedding: Where are the good latents?,
  2020{\natexlab{b}}.

\bibitem[Zhu et~al.(2021)Zhu, Abdal, Femiani, and Wonka]{zhu2021barbershop}
Peihao Zhu, Rameen Abdal, John Femiani, and Peter Wonka.
\newblock Barbershop: Gan-based image compositing using segmentation masks,
  2021.

\end{thebibliography}
\bibliographystyle{iclr2022_conference}

\FloatBarrier
\appendix
\section{Appendix: Additional Results}

\paragraph{Visual Evaluation of Style Transfer.}

\begin{figure*}[th]
    \centering
    \includegraphics[width=\linewidth]{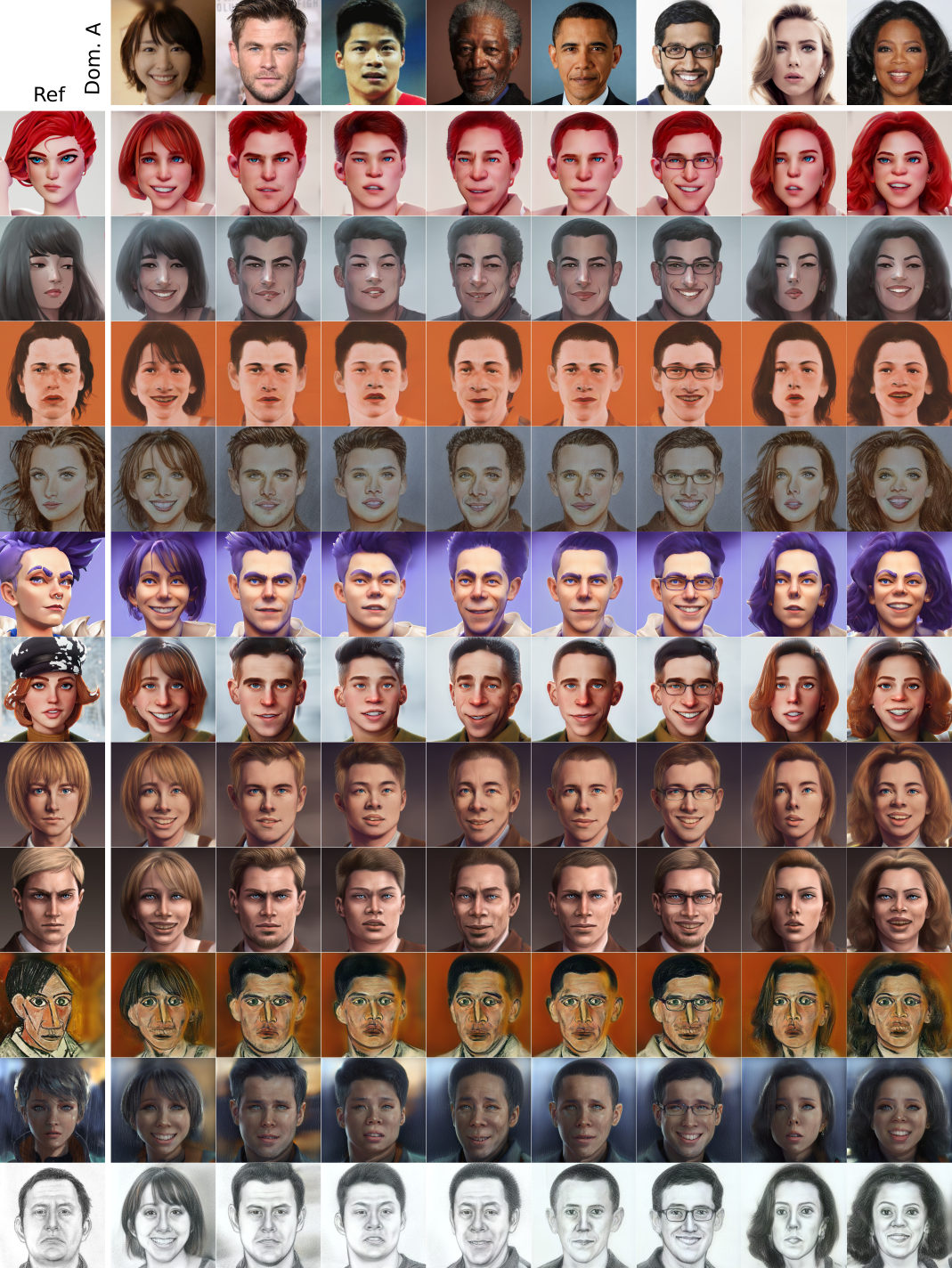}
    \caption{Style transfer results obtained by our method after style interpolation in domain $B$. The top row represents the real images embedded in the latent space of $G_A$ (domain $A$) whose latent codes are then used by $G_B$ (domain $B$). The first column represents the reference images $I_B$ which are input to our domain adaptation framework.}
    \label{fig:Basic_style_transfer}
\end{figure*}

\begin{figure*}[th]
    \centering
    \includegraphics[width=\linewidth]{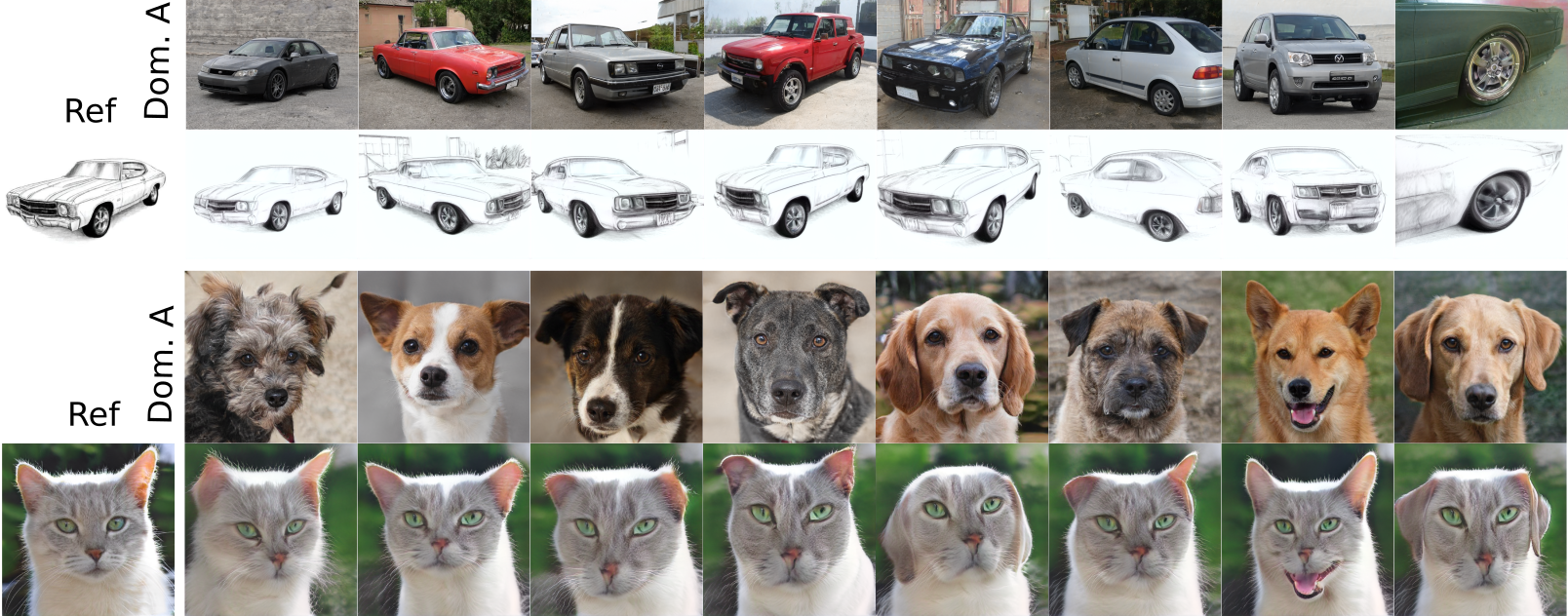}
    \caption{Our domain transfer results on cars and dogs. The structure of rows and columns is the same as in Fig.~\ref{fig:Basic_style_transfer}.  }
    \label{fig:Car_dataset}
\end{figure*}

We provide additional visual evaluation of the results by transferring style to existing photographs embedded into the StyleGAN latent space. In Fig.~\ref{fig:Basic_style_transfer} we show results for faces. The input photographs are in the top row and the style images in the first column. We can see that the results take on the style of the reference image, even though the reference image is far outside the original GAN's latent space. Also, we can notice that overfitting is successfully limited, as each result maintains several important aspects of the input image.
In Fig.~\ref{fig:Car_dataset} we show results for cars and dogs on the same task. This shows that our method is consistent across different StyleGAN objects/datasets.

\paragraph{Controlling the Style Gap.}

\begin{figure*}[th]
    \centering
    \includegraphics[width=\linewidth]{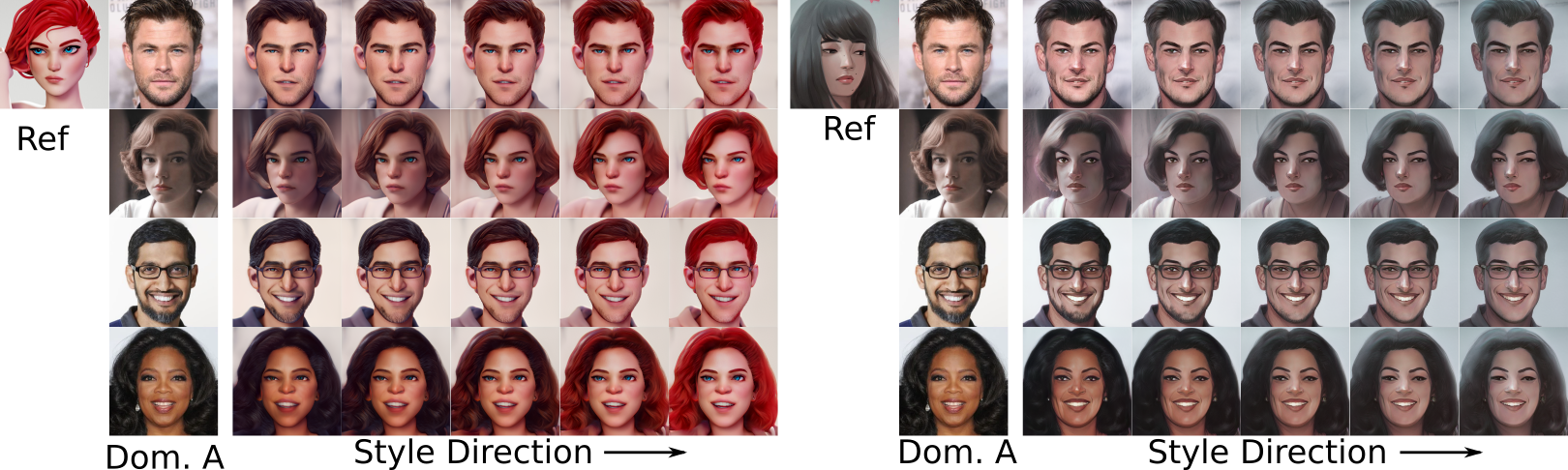}
    \caption{Style interpolation results achieved by our framework. Unlike the competing methods, our method has an explicit control over the styles in the domain $B$. Each sub figure shows a reference image and images embedded in domain $A$. Notice that we can control the amount of variation in style depending on a parameter $\alpha$ that can be specified by a user.}
    \label{fig:Style_interpolation}
\end{figure*}

Our method provides a way to control the domain gap between the domain $A$ and domain $B$ by explicitly controlling  the style of the images sampled from or embedded in domain $A$. 
Fig.~\ref{fig:Style_interpolation} shows that we can control the degree to which  style from the reference image is preserved by increasing the style-mixing parameter $\alpha$, which is not possible with any of the competing methods.
This gives users more control over  content generation and editing.

\begin{figure*}[th]
    \centering
    \includegraphics[width=\linewidth]{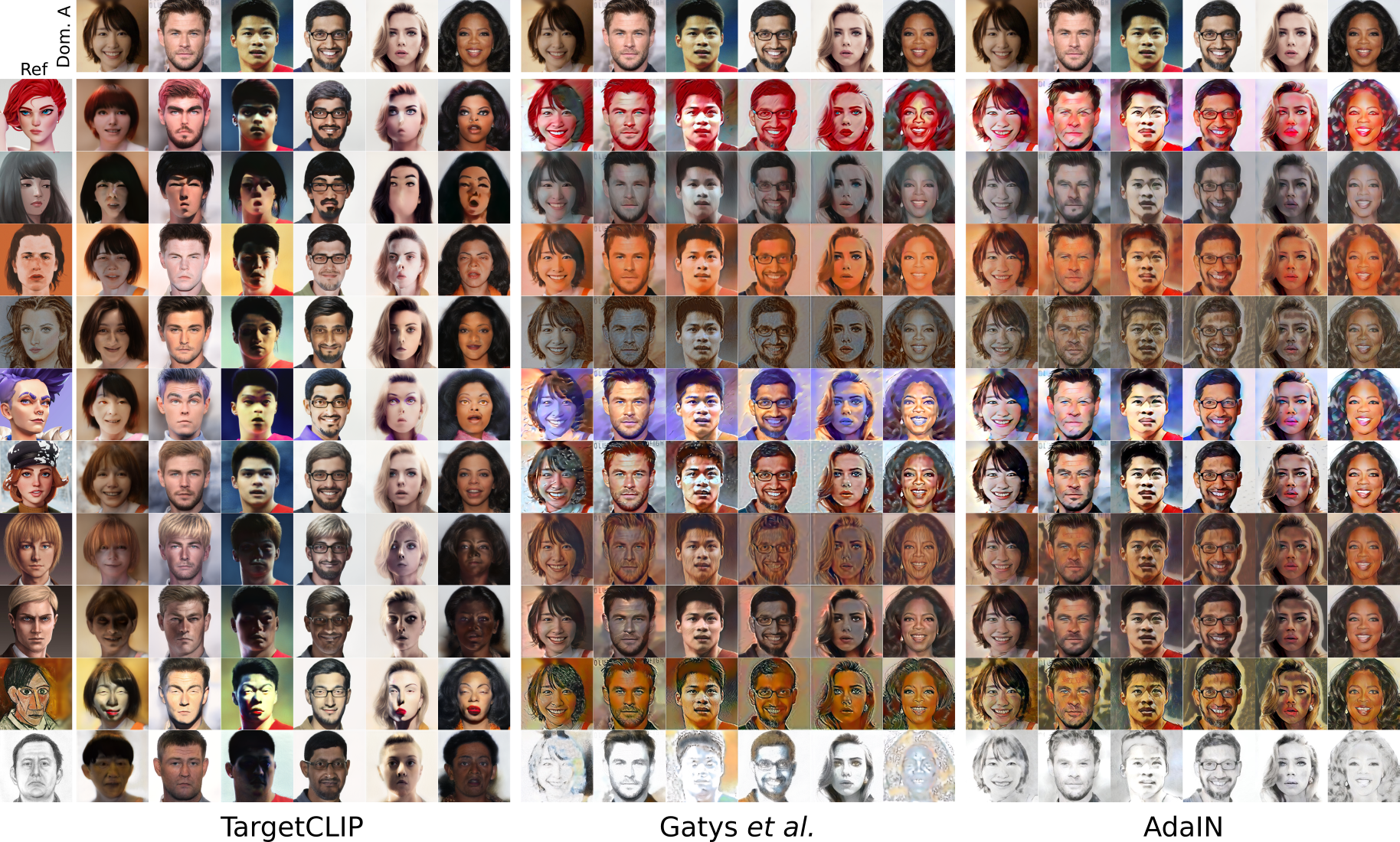}
    \caption{Additional omparisons with other baseline methods including the concurrent method TargetCLIP~\citep{chefer2021targetclip} as well as two lower-resolution methods from \citet{gatys2016image} and AdaIN~\citep{Adain2017}. One-shot reference images from domain $B$ are shown in the left column. Each image is the result of transferring the image in the top row into the new domain. Compare these images to our method in Fig.~\ref{fig:Attributes_control}, our proposed approach has fewer artifacts while preserving the identity of the image in domain $A$.}    \label{fig:other_methods}
\end{figure*}

\paragraph{Additional Comparison}
In addition to our comparison with StyleGAN-NADA~\citep{gal2021stylegannada} and few-shot domain adaptation~\citep{ojha2021fewshot}, we compare against three additional methods in Fig.~\ref{fig:other_methods}. These include one concurrently developed method called TargetCLIP~\citep{chefer2021targetclip} as well as two other methods that work on lower resolution images for one-shot domain transfer.  These are the method of Gatys \textit{et. al}~\citet{gatys2016image} and the the AdaIN approach~\citep{Adain2017}. Our visual results compare favorably against the new methods in Fig.~\ref{fig:other_methods} with respect to preserving the identity of the original image while also generating images that belong to the new domain. 

\begin{figure*}[th]
    \centering
    \includegraphics[width=\linewidth]{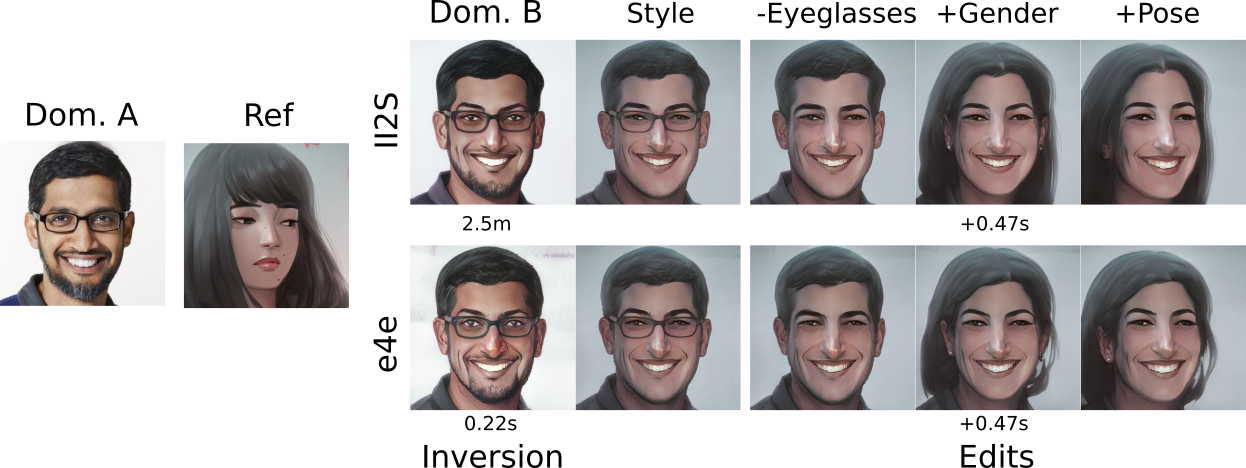}
    \caption{Comparison domain-transfer and editing using II2S vs e4e. The new GAN is always trained using II2S, but once training is complete, e4e can be used to transfer images into the new domain.
    II2S takes 2.5 minutes to embed the image, while e4e needs about 0.22 seconds. StyleFlow editing takes 0.47 seconds, and StyleGAN image generation takes about 0.34 seconds.}
    \label{fig:compareII2SvsE4E}
\end{figure*}

\paragraph{Inference and Editing Time}
Our proposed approach uses II2S for training and inference and StyleFlow~\citep{10.1145/3447648} for editing in the new domain.  GAN inversion using II2S on HD ($1024\times1024$) images takes 150 seconds on average, and each latent-code edit operation takes 0.47 seconds.  Generating the images afterwards takes an addition 0.34 seconds. Note that the run-time is dominated by GAN -inversion using II2S, however as we show in Fig.~\ref{fig:compareII2SvsE4E} once training is completed, we can use other GAN inversion methods to accomplish the edits. With e4e~\citep{tov2021designing} inversion is only 0.22 seconds and the entire process of inversion, editing, and generating the edited image can be accomplished in approximately one second.

\end{document}